# OnTarget: An Electronic Archery Scoring System

L.A. Danielescu
Department of Computer Science
University of Arizona
laviniad@cs.arizona.edu

## Abstract

There are several challenges in creating an electronic archery scoring system using computer vision techniques. Variability of light, reconstruction of the target from several images, variability of target configuration, and filtering noise were significant challenges during the creation of this scoring system. This paper discusses the approach used to determine where an arrow hits a target, for any possible single or set of targets and provides an algorithm that balances the difficulty of robust arrow detection while retaining the required accuracy.

## 1 INTRODUCTION

Currently, the structure of archery competitions does not allow for interaction between the spectators and the competitors. Scoring is done manually and is not posted until the end of the competition, making it impossible for the spectators to know each competitor's current score. To alleviate this problem, OnTarget, an electronic scoring system, was developed. OnTarget scores each player's shots, keeping track of and displaying players' scores and rankings to the spectators, making the competition more interactive.

The software is written in C++ using OpenCV, an open source image processing library developed by Intel. Two cameras connected to a frame around the target capture the images. The cameras are about 7" away from the edge of the target board and about 4" in front of the target (Figure 1).

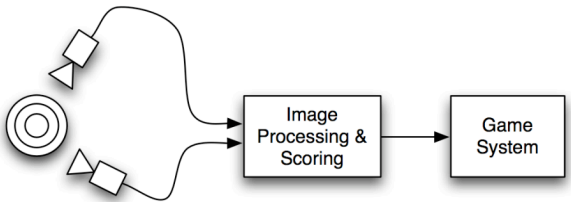

Figure 1: Setup of the target and cameras and flow of information.

OnTarget uses numerous computer vision techniques to detect the target and the location of the arrow. Ellipse fitting is used to detect the location of the target and then create a mask that can be used to filter out all extraneous noise in the image. Next, edge detection is used to detect individual rings on the target. Morphological operators are used to erode and dilate the image to remove some of the excess noise inside the target itself [6, 8].

To find the arrow within an image, the difference between the current and previous images is found. This provides a mask and from there a structuring element is used to find the arrow. From this, OnTarget can calculate the collision area between the arrow and the target, and can properly place the arrow within the rings that are identified in the edge detection step.

Section 2 covers background work that could be used in this type of application. Section 3 discusses the algorithm for recognizing the target within a given image, while section 4 presents the algorithm for detecting and scoring an arrow. Section 5 is the conclusion and section 6 discusses future research. Section 7 is the acknowledgements for this project.

## 2 BACKGROUND WORK

Although image processing has not yet been used in electronic archery scoring systems, there are many computer vision techniques that can be used for this particular application.

Several noise removal methods, such as those developed by Hirani et al. [7] and Quian et al. [9] can be used to increase the accuracy of the other methods.

Illumination cones can be used for variability of light, for faces [5] as well as finding the set of images of an object in all possible light conditions [3].

Incomplete edge detection is also a challenge in this application because it decreases the effectiveness of scoring the arrow. This problem, along with a potential solution, is discussed in Elder [4].

## 3 RECOGNIZING THE TARGET

Since scoring the arrow is not dependent upon the ground or the other scenery that will appear in the image, it is best to filter these out. To do this, the location of the target within the image needs to be found.

Several assumptions are made to achieve this in the current system. First, it is assumed that the target will have a white background so that it stands out. It is also assumed that there will be either a difference between the colors of the rings or edges between the rings that differentiate them, and that there will always be an X in the center of the target, which is traditionally the case in archery tournaments.

To begin, the user of OnTarget is asked to calibrate the cameras. After this, OnTarget takes pictures of the initial target (Figure 2). Then, the user must click on the center of each target for each image taken in order for OnTarget to register the target. Next, OnTarget finds all of the targets in the initial images. This is done through the use of edge detection and ellipse fitting. From these methods, OnTarget creates a mask that allows for the removal of anything outside of the targets.

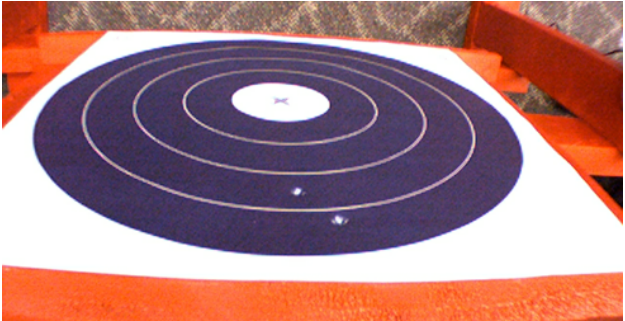
Figure 2: Original image taken by one of the cameras.

## 3.1 Edge Detection

OnTarget uses the canny edge detection algorithm provided in the OpenCV library. This algorithm is used to find the rings of the target, and therefore the regions that are scored differently (Figure 3).

Currently, since the edge detection does not always find the entire ring, accuracy is decreased and some arrows may not be scored correctly. Instead, the location of the nearest rings must be approximated.

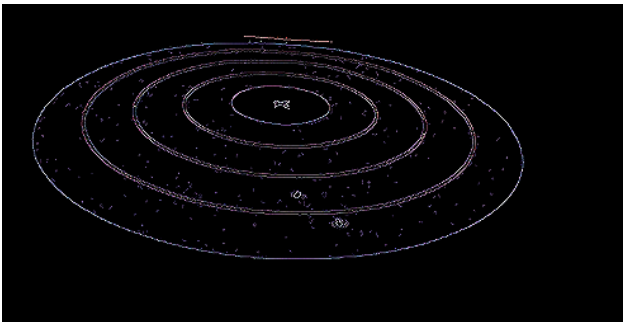
Figure 3: Edge detection on the target. Noise is not yet removed.

## 3.2 Removing Noise

Several different methods are used to attempt to remove excess noise within the area of the target. One of these is a bilateral smoothing algorithm incorporated in OpenCV. The image is also eroded and dilated once to remove any other small noise.

Noise removal is necessary to improve the accuracy of the scoring algorithm, specifically in the case of the contour detection. However, not all of the noise is removed in order to preserve image quality.

## 3.3 Ellipse Fitting and Creating the Mask

Ellipse fitting is used to find an overall mask of the target to block out the target's surrounding area (Figure 4). This mask can then be used to ignore anything in the image that is not relevant to scoring the arrow.

The ellipses are created using the contours of the edge detection. For all the contours found, OnTarget then checks to make sure the center is within 100 pixels of the user identified target center. This is done to eliminate any extraneous ellipses. The ellipse's height and width are expanded by 50 pixels each before it is drawn in order to account for arrows that may fall at the edge of the target.

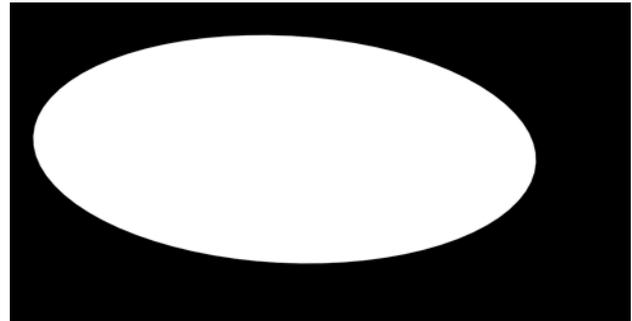
Figure 4: Mask of the target.

## 4 DETECTING AND SCORING AN ARROW

To detect and score the arrow OnTarget must first find its general location within the image. This is done by taking an image differential. Then, the arrow is extracted from the remaining noise using a vertical structuring element. From this the base of the arrow can be found.

In archery, if an arrow falls in two rings, it is scored based on the innermost ring. As such, the location of the point on the base of the arrow that is closest to the center of the target is found and scored accordingly.

The point value of that ring is then added to the current player's score.

## 4.1 Finding the Difference between Images

To find the location of the arrow on the image, OnTarget computes the difference between the previous and current images based on a threshold value. This is done by calculating the sum of the difference of the individual channels in RGB space (Figure 5). RGB space is used to preserve as much information about the different colors as possible.

After finding the general location of the arrow, the exact locations of the arrow's edges need to be found.

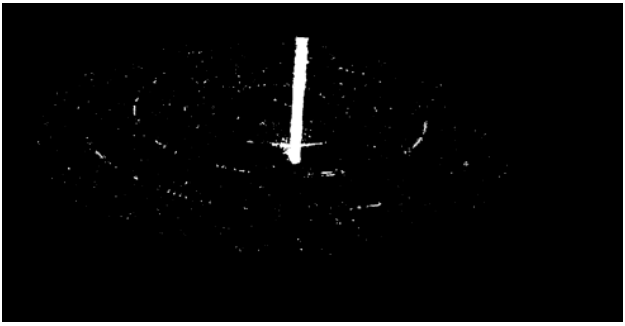
Figure 5: Difference between the image of just the target and the image with the arrow.

## 4.2 Finding the edges of the arrow

To find the edges of the arrow, OnTarget uses a vertical rectangular structuring element to extract the arrow from the surrounding area. This is done because there is still noise in the image and eroding and dilating the image is not an option due to the loss of precision. From the extracted arrow image (Figure 6), the closest point to the center of the target is found.

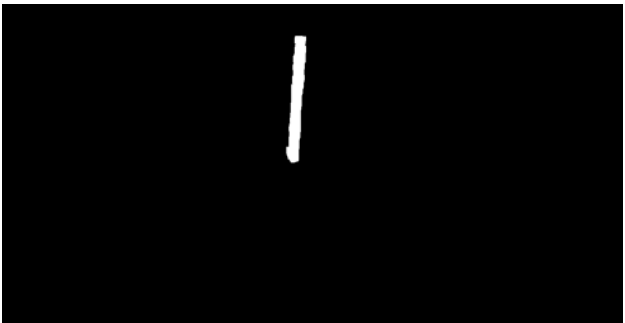
Figure 6: Extracted arrow from the structuring element.

## 4.3 Scoring the arrow

Once point that the score is going to be based on is found, OnTarget needs to find which ring this point falls into. For this step, a mask of each ring is created. This is done by running an ellipse fitting algorithm based on contours of the previously detected edges. Any ellipses found that are not large enough to be a ring within the target are discarded. All others are saved as masks for each ring of every target on the screen.

Then every mask found is applied in order of decreasing size and a check whether the point is still within that mask is performed. If it is not, the point falls within the previous ring that was checked and the arrow can be properly scored (Figure 7 and 8).

Since the most accurate scoring is desired and one camera may be able to provide a more accurate score than the other camera, both camera images are run through the same process. The highest score is then added to the player's current score.

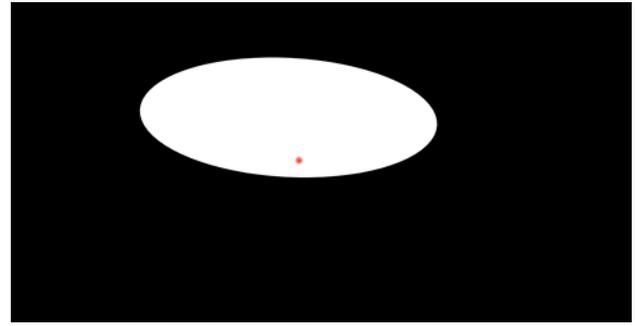
Figure 7: Mask of the ring that still encompasses the point of the arrow.

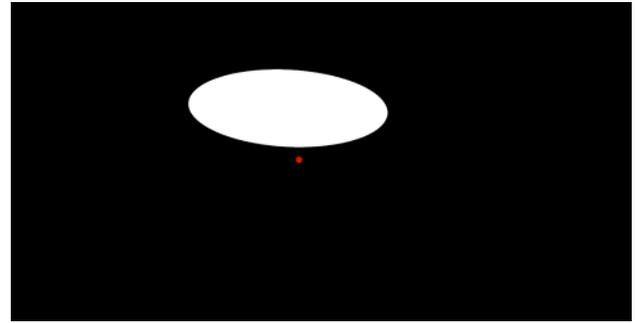
Figure 8: Next smallest mask. The point being scored is outside this mask.

## 4.4 Variable Lighting

Variable lighting causes large changes between images and can blur the image so that the edges and rings are not recognized.

Since large changes between images should not affect the quality of the scoring algorithm, the difference between the two most recent images is found. It is expected that there will not be large changes in lighting between shots, and instead will have greater changes in lighting throughout the competition. This allows for a larger variability of light in the span of a competition without greatly affecting the scoring.

## 5 RESULTS

Out of 16 scored arrows, an overall accuracy of 87.5% was obtained (See Figure 9). The arrows scored incorrectly were due to poor calibration on the side camera. By addressing the miscalibration of the cameras, a minimum accuracy of 93.75% would easily be achieved. At least one of the incorrectly scored arrows would be correctly scored by the side camera if poor calibration were not a problem.

|  | Correct | Incorrect | Total | Accuracy |
|---|---|---|---|---|
| **Side camera** | 11 | 5 | 16 | 68.75% |
| **Top camera** | 14 | 2 | 16 | 87.50% |
| **Overall** | 14 | 2 | 16 | 87.50% |

Figure 9: Results of running 16 image sets through OnTarget.

Unfortunately, there are several factors that limit the baseline of comparison between this method and a human judge. One limiting factor is the way in which a human judge scores an arrow. When a judge scores an arrow, he will walk up and look at the shot after all shots for that player have been taken. Our system, on the other hand, provides real time scoring. Also, according to Tom Green, Chair of USA Archery's Officials and Rules Committee, judges only score about 30 out of 18,000 shots in a competition. This means that the maximum possible error for a human judge would be .17%.

# 6 CONCLUSION

This method shows that, given certain constraints and assumptions, this type of system can be efficiently implemented in archery tournaments to increase interaction between the competitors and spectators. However, changes do need to be made to obtain accuracy closer to that of a human judge.

These changes include decreasing the sensitivity of the camera calibration, using a more accurate ellipse detection algorithm, addressing light variation, and adding a probabilistic edge completion algorithm. These changes, along with the current minimum resolution of 1280 x 960, would provide the required accuracy for competitions.

A decrease in the sensitivity of the calibration can easily be achieved by using better cameras. The cameras used in this setup did not focus well on the target. Addressing light variability would also decrease the sensitivity of the calibration. Light conditions can easily blur or brighten the images taken by the cameras. Removing these lighting problems is necessary to detect the target within the image.

A more efficient ellipse detection algorithm, such as those developed by Aguado et al. or Xie et al. should be used to create all masks. This will eliminate the high amount of ellipses that were detected which were not part of the target. Therefore, the target will be easier to detect accurately.

The canny edge detection algorithm used to find the contours of the ellipse and in several other steps produces incomplete edges with respect to some of the innermost rings. To detect all rings of the target accurately, a probabilistic edge completion algorithm should be added. This will also aid in ellipse detection.

Although this algorithm was developed for archery competitions, it can also be extended to gun competitions. A more simplistic version of this algorithm can also be used for any game that can be translated into a 2D scoring plane. One such example is in soccer to detect where a soccer ball enters the goal and measure its distance in relation to the goal keeper.

# 7 FUTURE RESEARCH

Future work includes transforming all of the camera images to one viewing plane to test whether or not this would increase scoring accuracy. Variability of light was also not stronger addressed in this research and automation is necessary if this were to be implemented in an industrial setting.

By transforming all of the camera images to one viewing plane, it is possible that a closer approximation of where the arrow has actually fallen can be found given the different camera angles that are provided. To do this the transform for all images to the plane must be found.

Figure 10 is an image of a target taken from a specific camera angle with lines drawn in for the triangulation needed to find the transform.

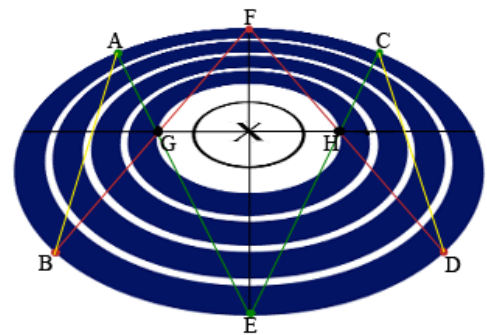

Figure 10: Finding the transform of the target.

Since the camera picks up the circular target as an ellipse, the goal is to find a transform between an ellipse and a circle and still maintain proportions.

In the above image, points A, B, C and D must be found to find the transformation matrix. Unfortunately the only points of reference that are currently available are points E and F, from the minor axis of the outermost ellipse of the target and the center of the X, from which points G and H can be found. These are used to find angles FEG, FEH, EFG, and EFH. From angle EFG, OnTarget can calculate point B by extending line segment FG out until it intersects the outermost edge of the target. Using this same process, points A, C, and D can be found.

To transform the ellipse into a circle, B and D would remain in the same place, but A and C would be moved so that the segments AB and CD remain the same length but are now parallel.

This work was done with one target of a specific size. Although most of the algorithm is scalable to multiple targets of varying sizes, there is still a need for minor changes to make the software completely scalable to the full complexity of competitive archery. The transformation matrix is one such change. These calculations assume there is only one target within the image and the calculations are based on the properties of the target. This algorithm therefore will not find the correct transformation for an image with multiple targets. Instead, a transformation needs to be found for the entire image rather than a per target basis.

Variable lighting conditions also need to be further addressed. Experimenting with an infrared filter might reduce many of the lighting problems that were encountered while trying to calibrate the cameras. If the image is too bright, the camera cannot pick up individual rings. Also, due to the low quality of the cameras, the white rings would show up as a rainbow of colors.

To further automate the software, there are several responsibilities that can be removed from the user. Instead of requiring the user to click on the center of the X within the image, the center of the target can instead be found by identifying that X automatically. Also, force feedback can be added to the backing of the target to signal the cameras to take the next image rather than having the user click a button after each arrow hits the target.

# 8 ACKNOWLEDGEMENTS

I would like to acknowledge Matt Poulter and Bryan Aull as additional group members on the OnTarget project, and also Jesse Lane, Graduate Student Mentor, and Jim Oliver, Faculty Mentor.

This research was performed at Iowa State University as part of the SPIRE-EIT research internship sponsored by the Human Computer Interaction Program and the Program for Women in Science and Engineering during the summer of 2007.

This material is based upon work supported by the National Science Foundation under Grant No. IIS-0552522. Any opinions, findings, and conclusions or recommendations expressed in this material are those of the author and do not necessarily reflect the views of the National Science Foundation.